# Global ECG Classification by Self-Operational Neural Networks with Feature Injection

Muhammad Uzair Zahid, Serkan Kiranyaz and Moncef Gabbouj

*Abstract*— **Objective:** Global (inter-patient) ECG classification for arrhythmia detection over Electrocardiogram (ECG) signal is a challenging task for both humans and machines. Automating this process with utmost accuracy is, therefore, highly desirable due to the advent of wearable ECG sensors. However, even with numerous deep learning approaches proposed recently, there is still a notable gap in the performance of global and patient-specific ECG classification performance. **Methods:** In this study, we propose a novel approach for inter-patient ECG classification using a compact 1D Self-ONN by exploiting morphological and timing information in heart cycles. We used 1D Self-ONN layers to automatically learn morphological representations from ECG data, enabling us to capture the shape of the ECG waveform around the R peaks. We further inject temporal features based on RR interval for timing characterization. The classification layers can thus benefit from both temporal and learned features for the final arrhythmia classification. **Results:** Using the MIT-BIH arrhythmia benchmark database, the proposed method achieves the highest classification performance ever achieved, i.e., 99.21% precision, 99.10% recall, and 99.15% F1-score for normal (N) segments; 82.19% precision, 82.50% recall, and 82.34% F1-score for the supra-ventricular ectopic beat (SVEBs); and finally, 94.41% precision, 96.10% recall, and 95.2% F1-score for the ventricular-ectopic beats (VEBs). **Significance:** As a pioneer application, the results show that compact and shallow 1D Self-ONNs with the feature injection can surpass all *state-of-the-art* deep models with a significant margin and with minimal computational complexity. **Conclusion:** This study has demonstrated that using a compact and superior network model, a global ECG classification can still be achieved with an elegant performance level even when no patient-specific information is used.

## I. Introduction

CARDIOVASCULAR diseases (CVDs) are responsible for 31% of deaths globally, according to the World Health Organization (WHO) [1]. It is crucial to detect CVDs as early as possible to begin effective treatment and medication. For cardiac arrhythmia detection, a variety of methods such as blood tests, stress tests, echocardiograms, and chest X-rays have been used. Still, ECGs are perhaps the most popular among clinicians. ECGs record the heart's electrical activity over time and can help diagnose many conditions, including premature ventricular contractions (PVCs or V rhythms) and supraventricular premature beats (SPBs or S rhythms). An experienced cardiologist can determine the presence of an arrhythmia, as an abnormality of heart rate or rhythm or a change in morphological pattern, by analyzing a recorded ECG signal. However, identifying and classifying arrhythmias can be an erroneous, labor-intensive, and subjective task even for cardiologists since it often requires considering each heartbeat of an ECG signal accumulated over hours or days. With the recent advances in various low-cost portable ECG devices [2], [3] such as chest straps and wristbands, the opportunities for self-monitoring and auto-diagnosis have increased. Therefore, it is highly desirable to have global (patient independent or inter-patient) and reliable ECG classification methods. However, robust and accurate classification of ECG signals still poses a challenge because among different patients or even for the same patient but under different temporal, psychological, and physical conditions, significant variations may occur in ECG signals' morphological and temporal/structural characteristics.

ECG-based arrhythmia classification is typically initiated with a peak detection/segmentation. This study does not discuss R-peak detection since highly accurate algorithms have already been proposed in the literature [4], [5]. The analysis and classification of ECG signals have been extensively studied throughout the last decades [6]–[9]. Generally, these works can be classified as intra-patient, inter-patient (global), and patient-specific [10]. In the intra-patient paradigm, datasets are divided into training and test subsets according to heartbeat labels. Therefore, beats from the same individual may appear both in training and evaluation subsets, making the evaluation process biased [11]. The classifiers usually produce over-optimistic results (in close vicinity of 100%) because the model learns the information specific to the patient during the training phase [11]–[13]. The classification performance declines due to inter-individual variability. Hence, morphological variations in ECG from different patients should be considered when building the model. Even for a healthy subject's (normal) ECG waveform, the shape of the QRS complex, P waves, and R–R intervals may differ from one beat to the next under various circumstances [14]. Chazal et al. [15] presented the inter-patient paradigm where training and testing heartbeats are collected from different patients' ECG recordings to adopt real-world scenarios. Some patients are reserved for the evaluation phase and beats from other patients are used to

Muhammad Uzair Zahid and Moncef Gabbouj are with the Department of Computing Sciences, Tampere University, 33100 Tampere, Finland (e-mail: muhammaduzair.zahid@tuni.fi; moncef.gabbouj@tuni.fi).
Serkan Kiranyaz is with the Department of Electrical Engineering, College of Engineering, Qatar University, Doha 2713, Qatar (e-mail: mkiranyaz@qu.edu.qa).

train the classifier so that classifier would exhibit a better generalization capability for new unseen patients. Most researchers have chosen to use another approach called the "patient-specific" paradigm, i.e., [9], [16]–[18] in which other patients and patient-specific beats of a new patient are jointly used to train the network. Although the patient-specific paradigm is superior to inter-patient paradigms in terms of performance, it requires cardiologists' labeling in advance for each (new) patient, which is cumbersome, subjective, and labor-intensive. Furthermore, a new network should always be trained from scratch or fine-tuned carefully to achieve the required generalization. Only then one can evaluate and test its reliability, all of which limit its clinical application. Especially, the training of these personalized models requires the collection of patient-specific arrhythmic data, which requires long-term monitoring or even may not exist. As the data volume increases, it becomes difficult or even impossible to manually label small chunks of data from all stored records. In a recent work [19], an adaptive patient-specific heartbeat classification model is proposed for diagnosing heart arrhythmias. A general classifier was first trained on the general population. Then, the weights in the lower part of the general classifier were retained using patient-dependent i-vectors and the weights in the upper part were randomized.

For several decades, feature engineering-based methods dominated ECG signal recognition. Studies [20]–[23] based on traditional signal processing and machine learning methodologies have not been successful in clinical settings. This is because there can be significant variations in the morphological characteristics and temporal/structural dynamics of ECG signals for different patients or even the same patient under varying physical, psychological, and temporal conditions. Such hand-crafted feature extraction may not capture the actual characteristics of each ECG beat variation for accurate classification. Therefore, their performance level varies significantly in large ECG datasets [24]. Moreover, extreme performance variations may occur due to increased noise levels, different ECG sensor types, inter-patient variations in ECG signals, and different arrhythmia prevalence between databases.

Mariano *et al.* [25] extracted features using both leads of ECG, wavelet transform, and RR intervals. The floating feature selection model was used to reduce the feature set, and finally, eight features were fed into the classifier. Can *et al.* [26] extracted morphological features using wavelet transform and dynamic features using RR intervals. A combination of these features is then fed into the SVM classifier. The authors reported an overall accuracy of 86.4% in the patient-specific evaluation. In another study [27], a weighted variant of the conditional random fields classifier (CRF) was used with L1 regularization and achieved an accuracy of 85%. Khorrmi *et al.* conducted a comparative study of feature extraction and classification methods. The Discrete Wavelet Transform (DWT), Continuous Wavelet Transform (CWT), and Discrete Cosine Transform (DCT) were compared to extract features. Similarly, a comparison between a multilayer perceptron (MLP) and a support vector machine (SVM) was presented as a classifier [28]. Karpagachelvi *et al.* combined discrete wavelet transform with high-order statistics and AR modeling to extract features, while extreme learning machines (ELMs) were used for classification [29]. For feature extraction, the authors used a vectorcardiogram-based ECG representation. Features were selected using the particle swarm optimization algorithm to feed into the SVM classifier [30].

The development of deep learning models has led to the widespread use of neural networks in many applications, including face detection, image denoising, image classification, and numerous one-dimensional signal processing. Recently, one-dimensional convolutional neural networks (1D-CNN) have also been extensively studied because of their speed and efficiency when managing complex tasks, as demonstrated by various applications involving signal processing [31], [32], motor fault detection [33], and advance warning system for cardiac arrhythmias [34]. In a study by Kiranyaz *et al.* [35], only three layers of a compact 1D CNN were used for patient-specific ECG classification. To train each personalized classifier, the authors used only the first 5-min section of the record and 245 common beats randomly selected from the train partition of the MIT-BIH dataset, following the AAMI recommendations [36].

Several *global* ECG classification methods [29], [37]–[41] based on deep CNN models have recently been proposed. They naturally have high complexity and require large volumes of labeled ECG data for training. In addition, because they require parallelized hardware to function, they cannot be directly implemented on low-power or mobile devices. Finally, most methods tested on unseen patients do not perform well in the inter-patient paradigm. As opposed to beat-wise classification, alternative deep learning approaches used ECG segments instead. In their study, Acharya *et al.* used two and five seconds of ECG data with a 10-layer CNN model [42]. The deep network architecture proposed by Li *et al.* consists of densely connected CNNs (DenseNet) further connected to gated recurrent units (GRU). 10-second ECG segments were analyzed, and the F1 score was only 61.25% for SVEB detection and 89.75% for VEB detection [43]. The results have shown that even with deep network models, especially the SVEB detection performance is relatively poor in general, which hinders their clinical usage.

Recent studies [44]–[49] have pointed out that CNNs are homogeneous networks with an ancient linear neuron model that originated in the 1950s (McCulloch-Pitts). The linear neuron model is a crude representation of biological neurons with specialized electrophysiological and biochemical properties in highly heterogeneous biological networks [29]. [44]–[49]Following this, Operational Neural Networks (ONNs) [44], [50] have been proposed to address such drawbacks. ONNs derived from Generalized Operational Perceptrons (GOPs) [44]–[49] are heterogeneous networks with a nonlinear neuron model, which permits them to learn highly complex and multimodal functions or spaces with minimal network complexity and training data. Studies [51]–[53] have proposed the latest ONN variant, Self-Organized ONNs (Self-ONNs), for various image processing and regression tasks.

In this study, we propose a novel inter-patient ECG classification approach to address the aforementioned issues. The proposed method is based on compact 1D Self-ONNs, with feature injection/fusion ability. In our method, normalized ECG signals are divided into 230 samples (639 ms) of fixed-duration frames using the R peak as a reference point. To accomplish a multi-scale representation, each frame is decomposed into the time-frequency domain using Discrete Wavelet Transform (DWT) at nine different scales to achieve the scale invariance. As arrhythmia affects not only the morphology of the heart cycle but also varies the timing of beat, four R-R interval-based features are extracted and injected into the Self-ONN model to enrich the learned features. We evaluate the proposed approach on the MIT-BIH database. Overall, the novel and significant contributions of this study can be enlisted as follows:

1. We developed a compact architecture with 1D Self-ONN layers for global ECG classification that significantly outperforms all *state-of-the-art* methods.
2. This is the first study that proposes Self-ONNs with feature injection to perform a joint classification in the same network.
3. A multi-scale approach using DWT is proposed to transform the raw ECG signal before feeding it to the network and thus achieve the scale invariance.
4. Finally, over the MIT-BIH benchmark dataset, we show that without changing or fine-tuning the model, the performance of our model remains the same for unseen patients despite the morphological variations.

The rest of the paper is organized as follows: Section II outlines the ECG datasets used in this study. The proposed approach is presented in Section III. In Section IV, the performance of the proposed system is evaluated over the MIT-BIH database using the standard performance metrics, and the results are compared with the recent *state-of-the-art* works. Finally, Section V concludes the paper and suggests topics for future research.

## II. 1D SELF-ORGANIZED OPERATIONAL NEURAL NETWORKS

Figure 1 shows 1D nodal operations of a CNN, ONN with fixed (static) nodal operators, and Self-ONN with generative neuron which can approximate any arbitrary nodal function, $\Psi$, (including the conventional types such as linear, exponential, Gaussian, or harmonic functions) for each kernel element of each connection. With such generation ability, obviously, Self-ONNs have the potential to achieve greater operational diversity and flexibility, allowing the optimal nodal operator function to be formed for each kernel element to maximize the learning performance. Another crucial advantage over conventional ONNs is that Self-ONNs do not use an operator set library or a prior search process to select the best nodal operator.

The $Q^{th}$ order truncated approximation, formally known as the *MacLaurin* polynomial, takes the form of the following finite summation:

$$\psi(x)^{(Q)} = \sum_{n=0}^{Q} \frac{\psi^{(n)}(0)}{n!} x^n \qquad (1)$$

The above formulation can approximate any function $\psi(x)$ sufficiently well near 0. When the activation function bounds the neuron's input feature maps in the vicinity of 0 (e.g., *tanh*) the formulation of (1) can be exploited to form a composite nodal operator where the power coefficients, $\frac{\psi^{(n)}(0)}{n!}$, can be the learned parameters of the network during training.

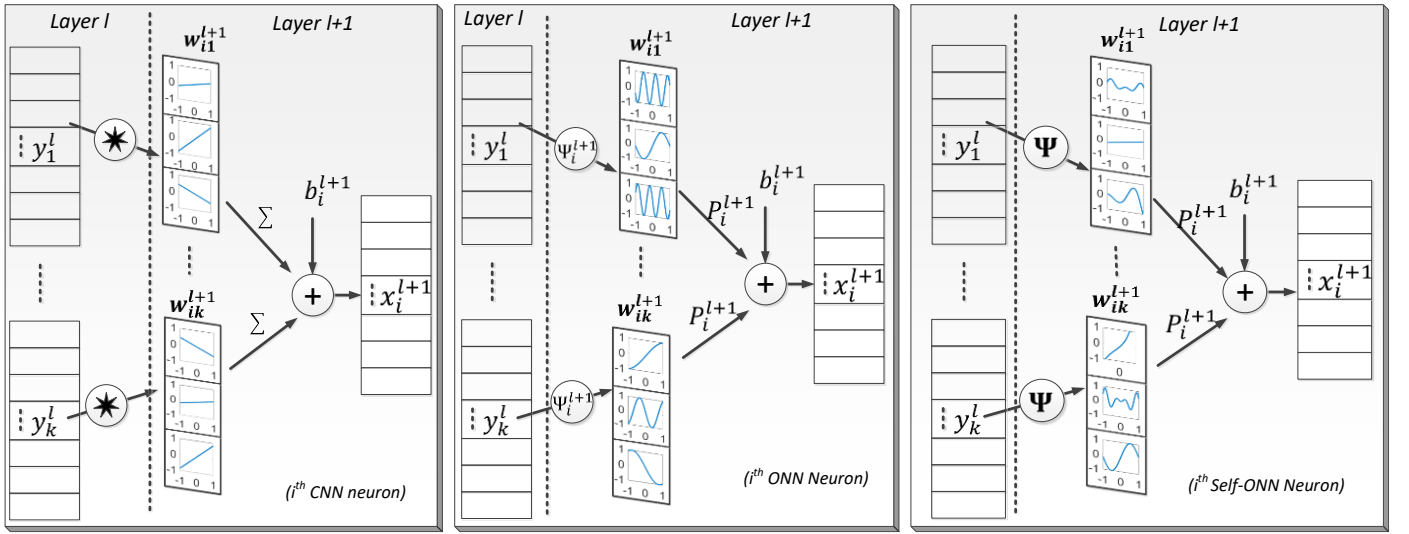

**Figure 1: An illustration of the 1D nodal operations with the 1D kernels of the $k^h$ CNN (left), ONN (middle), and Self ONN (right) neurons at layer $l$** [54]

It was shown in [52] that the nodal operator of the $k^{th}$ generative neuron in the $l^{th}$ layer can take the following general form:

$$\widetilde{\psi_k^l}\left(w_{ik}^{l(Q)}(r), y_i^{l-1}(m+r)\right)$$
$$= \sum_{q=1}^{Q} w_{ik}^{l(Q)}(r,q) \left(y_i^{l-1}(m+r)\right)^q \quad (2)$$

$$\widetilde{x_{ik}^l}(m) = \sum_{r=0}^{K-1} \sum_{q=1}^{Q} w_{ik}^{l(Q)}(r,q) \left(y_i^{l-1}(m+r)\right)^q \quad (3)$$

where K is the size of the 1D kernel of the $i^{th}$ neuron at layer $l$. One can simplify Eq. (3) as follows:

$$\widetilde{x_{ik}^l} = \sum_{q=1}^{Q} Conv1D\left(w_{ik}^{l(Q)}, (y_i^{l-1})^q\right) \quad (4)$$

Hence, the formulation can be accomplished by applying Q of 1D convolution operations. Finally, the output of this neuron can be formulated as follows:

$$x_k^l = b_k^l + \sum_{i=0}^{N_{l-1}} \widetilde{x_{ik}^l} \quad (5)$$

where $b_k^l$ is the bias associated with this neuron. The $0th$ order term, $q = 0$, the DC bias, is omitted as its additive effect can be compensated by the learnable bias parameter of the neuron. With the $Q = 1$ setting, a generative neuron reduces back to a convolutional neuron.

The raw-vectorized formulations of the forward propagation, and detailed formulations of the Back-Propagation (BP) training in raw-vectorized form can be found in [52] and [53].

## III. METHODOLOGY

The proposed global ECG classification approach is illustrated in Figure 2. The single-channel raw ECG signal is the first unit normalized and partitioned into the segment of 230 samples. Then continuous wavelet transform is employed to convert the 1-D ECG beat into a nine-channel time-frequency beat representation (9x230) which is fed into the proposed 1D Self-ONN model. The temporal features are then injected into the Self-ONN classifier to accomplish the final classification of the ECG beat.

### A. Problem formulation

In general, abnormalities in the ECG signals can be linked to two main aspects: ECG beat morphology (morphological variations) and the time interval between ECG beats (temporal variations). As illustrated in Figure 4, the top figure shows a premature beat (temporal variance between R peaks), and the bottom figure shows ill-shaped QRS complexes (morphological variance). The morphology of each heartbeat plays a vital role in classifying arrhythmia. Moreover, the timing or location of the heartbeat is also a crucial feature. Segment-based classification can detect abnormalities without looking at timing data explicitly since the input is based on multiple beats or cardiac cycles. On the other hand, when a continuous ECG signal is divided into frames, each with a single beat, this will cause the loss of temporal information. It is relatively difficult to distinguish a beat from another by just examining its morphology. As can be seen in the figure, a regular N-beat and S-beat look similar. However, exploiting the beat location makes it easy to distinguish between them. This is the main reason for injecting the R-R-based features into the feature representation learned from Self-ONN layers to make the final classification.

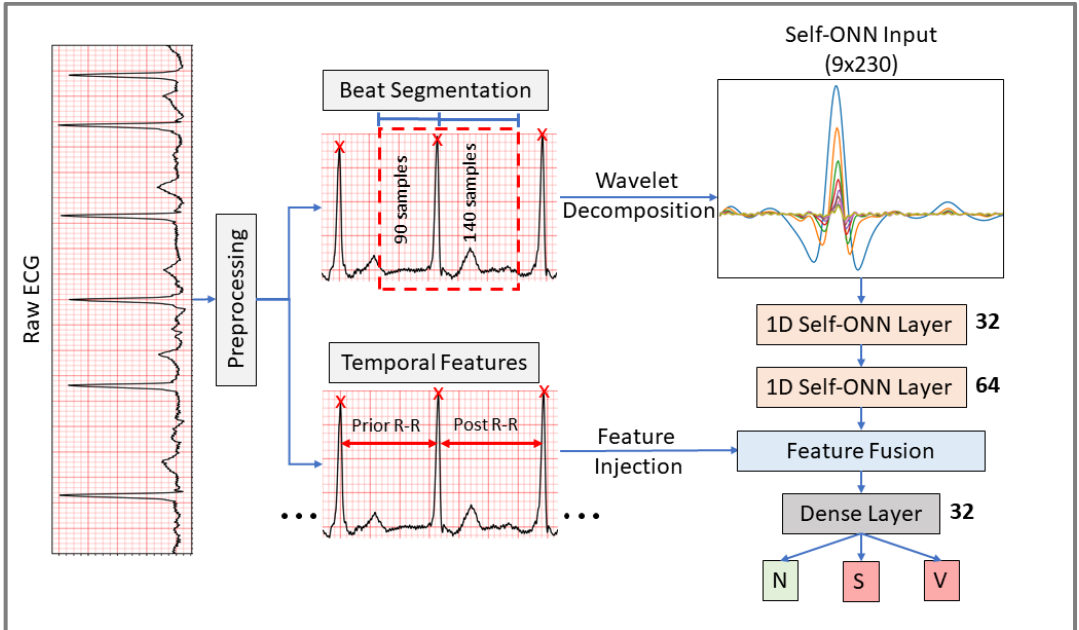

**Figure 2: Block diagram of the proposed approach and model architecture for classification of ECG signals.**

## B. Data Processing

**Beat Segmentation:** Each heartbeat's morphology is crucial to classifying arrhythmias. In some studies, the R peak is used as a center to segment the ECG signals. However, this is not a good approach in practice as the QT interval is approximately double the PR interval in duration. We do not need the morphological information before the P wave as it falls within the previous heart cycle boundary. PR intervals are generally between 0.12 and 0.20 seconds in duration and extend from the onset of the P wave to the beginning of the QRS complex. The QRS complex usually lasts between 0.06 and 0.10 seconds. The QT interval can range from 0.20 to 0.44 seconds depending upon heart rate. To avoid any information loss, the upper bound of each duration is considered. A detailed description of one ECG cycle and its waves are presented in Figure 4. By keeping the upper bound of these intervals and sampling frequency of 360Hz into consideration, the ECG signal from a single channel was segmented into heartbeats using the R peak as a reference point, taking 250msec before and 390msec after the peak.

**Temporal Features:** In order to derive information about the timing of the ECG beat or, more precisely, about the temporal variation, we extracted four widely used R-R-based features, i.e., prior R-R interval, post R-R interval, a ratio of prior to post R-R interval and average R-R interval over ±10 seconds from the current beat.

**Multi-scale ECG representation:** Multi-scale ECG representation plays an important role in the classification of ECG signals. Any feature that can be used to discriminate abnormal beats from normal ones can be revealed in different frequency scales. Such features can represent the time-frequency characteristics of the raw ECG signal. Some prior studies have investigated several methods for transforming ECG signals into different scales before feeding them to the classifier network. Among them, the DWT is considered to be the most efficient for processing ECG signals. When using DWT, ECG information can be retrieved both in the frequency and time domains, which is far superior to the DFT, which can only analyze ECG information in the frequency domain. This study applies DWT based on the Ricker (Mexican-hat) wavelet to transform ECG beats at nine different scales to generate one-dimensional DWT patterns in each scale. The overall multi-scale representation can be viewed as a 9-channel representation of the original signal in different subband frequencies. Figure 3 shows examples of DWT transformation patterns from normal and arrhythmic ECG signals. Over an ECG segment $x(t)$, its DWT with respect to a given mother wavelet $\psi$ is defined as follows:

$$W_x(a,b) = \frac{1}{\sqrt{a}} \int_{-\infty}^{\infty} x(t)\psi\left(\frac{t-b}{a}\right)dt \quad (6)$$

where $a$ is a scale parameter and $b$ is a translation parameter. The scale can be converted to frequency by

$$F = \frac{F_c \cdot a}{f_s} \quad (7)$$

where $F_c$ is the center frequency of the wavelet basis, $f_s$ is the sampling frequency of the signal. In this study, we used a specific set of 9 scales. The corresponding band frequencies of the scales range from 10 Hz to 90 Hz with a gap of 10 Hz.

Finally, we used the Ricker wavelet basis defined as,

$$\psi(t) = \frac{2}{\sqrt{3} \cdot \sqrt[4]{\pi}} e^{\frac{-t}{2}}(1-t^2) \quad (8)$$

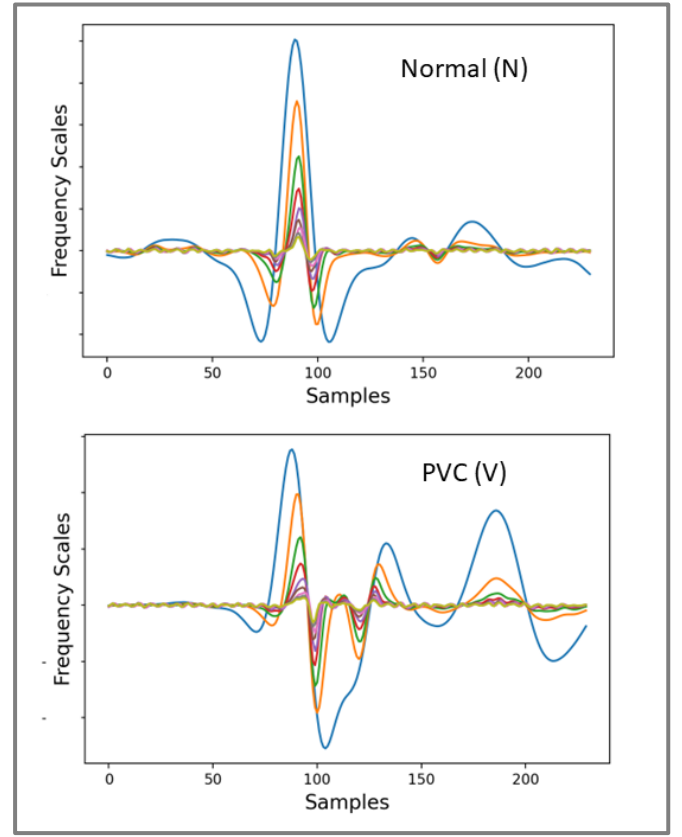

**Figure 3: Transformation patterns of a normal heartbeat (top) and an abnormal heartbeat due to premature ventricular contractions (bottom).**

## C. Data Augmentation

To achieve a more balanced distribution of classes so that underrepresented arrhythmias would become more prominent, we augmented the arrhythmia beats instead of excluding the majority class samples during the training of the classification model. Data augmentation is crucial to achieving the network's robustness and invariance with limited or imbalanced training samples across different classes. We found that arrhythmia beats (S and V beats) are far less frequent than the normal beats in the MIT-BIH dataset. As described in [5], we generated augmented arrhythmic rhythms from the 20-second ECG segments containing one or more arrhythmia beats by adding baseline wander and motion artifacts from the Noise Stress Test Database [55].

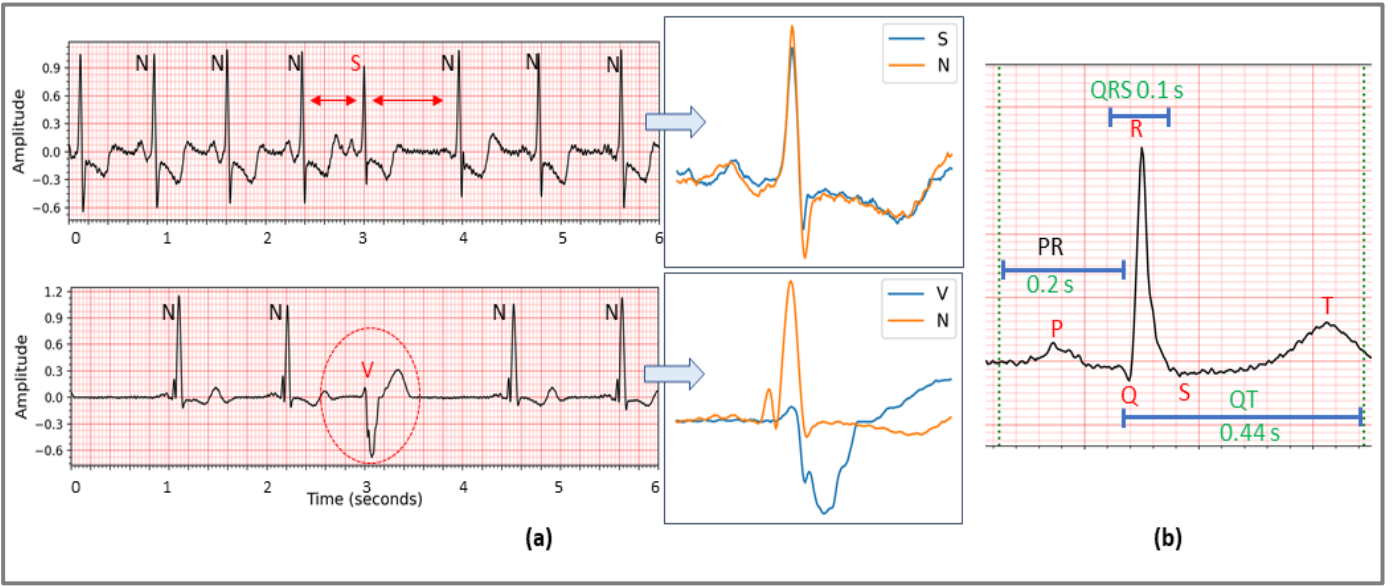

**Figure 4:** (a) On top, the S beat exhibits temporal variation. In contrast, the beat morphology of successive S and N beats is almost identical. We can observe variation in the subsequent N and V beat morphology in the bottom plot. (b) ECG cardiac cycle and intervals between different waves.

**Table I: Mapping the MIT-BIH Arrhythmia Database Heartbeat Types to the AAMI Heartbeat Classes [23].**

| AAMI Beat Class | MIT-BIH Normal/Arrhythmia types |
|---|---|
| Non-ectopic beats (N) | Normal (N) |
| | Right bundle branch block beats (RBBB) |
| | Atrial escape beats (e) |
| | Left bundle branch block (LBBB) |
| | Nodal (junctional) escape beats (j) |
| Supraventricular ectopic beats (S) | Aberrated atrial premature beats (a) |
| | Atrial premature contraction (A) |
| | Supraventricular premature beats (S) |
| | Nodal (junctional) premature beats (J) |
| Ventricular ectopic beats (V) | Premature ventricular contraction (PVC) |
| | Ventricular escape beats (E) |
| | Ventricular |
| | flutter wave (!) |
| Fusion beats (F) | Fusion of ventricular and normal beat (F) |
| Unknown beats (Q) | Paced beats (/) |
| | Fusion of paced and normal beats (f) |
| | Unclassifiable beats (Q) |

*D. Network Architecture*

We have implemented the 1D Self-ONN model as illustrated in Figure 2. This model, in brief, consists of two operational layers to extract learned features, fused with the injected temporal features, and two dense layers to analyze the combined features for classification. The first operational layer has 32 neurons with a filter size of 1x3, followed by a max-pooling layer of size 1x7. The second operational layer has 64 neurons with a filter size of 1x3, followed by an adaptive max-pooling of size 1x1, which applies the 1D adaptive max pooling over an input signal composed of several input planes. Both Self-ONN layers are followed by batch normalization and a hyperbolic tangent activation function (*tanh*). The output feature maps of operational layers are concatenated with the injected temporal features and fed into the dense layer where there are 32 neurons followed by rectified linear activation function (ReLu). The network's output layer size is 3, which computes the class score corresponding to each ECG class.

## IV. EXPERIMENTAL RESULTS

In this section, we first present the benchmark dataset, MIT-BIH, used for training and evaluation of the proposed approach. Then the metrics used for evaluating the proposed approach will be presented. Next, we will present a comprehensive set of experiments and comparative evaluations against the current *state-of-the-art* methods from the literature over the MIT-BIH dataset.

*A. Dataset*

As the gold-standard benchmark dataset, the MIT-BIH arrhythmia dataset [56] was used for performance evaluation in this study. Each recording on the MIT/BIH dataset is about a 30-minute duration and includes two-channel ECG signals. Each record is taken from the 24-hour ECG signals of 47 subjects. Every ECG record is preprocessed using band-pass filtering at 0.1-100 Hz and then sampled at 360 Hz. Independent experts have annotated both timing and beat class information in the database. Advancement of Medical Instrumentation (AAMI) classifies heartbeats in this database into 15 classes. Further, it divides them into five categories, which are normal (N), supraventricular ectopic beats (SVEB), ventricular ectopic beats (VEB), fusion beats (F) and unknown beats (Q), as shown in Table I. While the MIT-BIH arrhythmia dataset is frequently used, few studies follow the AAMI class division scheme and a more realistic evaluation protocol (inter-patient paradigm).

Table II: Each partition's records and the number of representative beats for each class.

| Partition | Patients | N | S | V |
|---|---|---|---|---|
| DS1 | 101, 106, 108, 109, 112, 114, 115, 116, 118, 119, 122, 124, 201, 203, 205, 207, 208, 209, 215, 220, 223, 230 | 45786 | 941 | 3784 |
| DS2 | 100, 103, 105, 11, 113, 117, 121, 123, 200, 202, 210, 212, 213, 214, 219, 221, 222, 228, 231, 232, 233, 234 | 44177 | 1834 | 3218 |
| Total | | 89963 | 2775 | 7002 |

A widely used data division method proposed by de Chazal *et al.* [11] is utilized to split the database in order to make a fair comparison with existing works. ECG recordings from 44 patients were divided into two datasets: DS1 and DS2, each containing ECG data from 22 recordings of approximately equal proportions of beat types. There are approximately 50,000 heartbeats in both partitions, including routine and complex arrhythmia recordings. The first dataset (DS1) was used to train and validate the classifier, while the second dataset (DS2) served as the basis for the final performance evaluation.

In Table II, ECG record partitions and the number of heartbeats for each class are presented. According to the AAMI recommended practice, we removed the four recordings (102, 104, 107, and 217) containing paced beats from the analysis because those patients were all wearing cardiac pacemakers that could potentially interfere with the analysis. Among the 44 ECG records from the MIT/BIH arrhythmia database, there are records with patient IDs in the range of 100 to 124 that reflect the common clinical ECG patterns. Other patient records with patient IDs ranging from 200 to 234 contain less common to very rare arrhythmia beats including ventricular, junctional, and supraventricular arrhythmias.

### B. Evaluation metrics

In this section, we present five of the most commonly used performance metrics to evaluate arrhythmia classification methods: accuracy (Acc), specificity (Spe), sensitivity (Se), positive predictive (Ppr), and F1-score. Majority class figures can significantly distort overall accuracy. As the classes for heartbeat types in the MIT-BIH database are highly imbalanced, the other four metrics are more relevant to comparing the methods.

$$Acc = \frac{TP + TN}{TP + FP + TN + FN} \times 100 \quad (9)$$

$$Sen\ (Recall) = \frac{TP}{TP + FN} \times 100 \quad (10)$$

$$Spe = \frac{TN}{TN + FP} \times 100 \quad (11)$$

$$Ppr\ (Precision) = \frac{TP}{TP + FP} \times 100 \quad (12)$$

$$F1 = \frac{2 \times Sen \times Ppr}{Sen + ppr} \times 100 \quad (13)$$

where TP is true positive, TN is true negative, FP is false positive and FN is false negative. As in the competing methods, we evaluated the classification performance for N, S, and V beats individually.

In addition, receiver operating characteristics (ROCs) were used to illustrate the diagnostic ability of a binary classification system with different thresholds of discrimination (specifically S and V beats). Due to the wide range of thresholds, ROC curves can provide comprehensive information regarding performance.

### C. Experimental setup

The proposed Self-ONN model is implemented using the FastONN library, a fast GPU-enabled library developed in Python and PyTorch to implement and train operational neural networks. The optimized PyTorch implementation of Self-ONNs is publically shared in [57]. The Adam optimizer is used with a learning rate (LR) of 0.01 and an LR scheduler, which drops by 0.1 every 10 epochs. Kaiming initializer was used to initialize the weights of the model. The model is trained for 35 epochs with a batch size of 128. Patient-wise, 5-folds cross-validation is used to train the model and tune the hyper-parameters. We used the cross-entropy loss as the objective function for training the network [2], which is then summed over all the samples in a mini-batch. The experiments were conducted on a computer equipped with an Intel Core i7-8750H, 16GB memory, 6 GB NVIDIA GeForce GTX 1060 graphics card, and a 2.21 GHz processor.

Additionally, the only parameter of the 1D Self-ONN (aside from the network configuration and the common hyperparameters) is the setting of Q (the order of the Taylor polynomial), which represents the degree of non-linearity for each neuron. When this value is set high, higher-order polynomials can be generated, resulting in a higher degree of nonlinearity, but at the expense of increasing the number of network parameters and complexity. In contrast, setting it too low will result in the opposite outcome. Setting it to Q=1 will result in Self-ONN being identical to a CNN, resulting in reduced learning and generalization performance. In order to achieve a balanced network, we choose Q=3 for all layers/neurons.

### D. Performance Evaluation

Two experiments were conducted so that we could compare the model's performance in real-world scenarios and demonstrate the robustness of the model. To begin with, we treated the ECG records in DS2 as unseen patient records and used those records as the test set, whereas DS1 was used for model training. Next, we will swap the training/evaluation sets, i.e., training on DS2 and evaluation on DS1. The three major classes (N, S and V) were considered in the experiments, while the other two classes, F and Q are ignored as in several studies [30]. In Table III, the confusion matrix is shown for all the records in both partitions of data (DS1 and DS2) of the MIT-BIH arrhythmia database using Q=3 in all Self-ONN layers. A more extensive and accurate comparison of performance evaluation is conducted by comparing the performance of the proposed system with the six existing algorithms, including the self-ONN network model, presented in [53]. Table IV presents the performance metrics of all methods. Several interesting observations can be made from the results in Table IV. First, for S-beat detection, sensitivity and positive predictivity rates are comparably lower than V-beat detection, while a high-specificity performance is achieved. The first and foremost reason for slightly worse performance in detecting S-beats as compared to V-beats is that the S class is underrepresented in the training data, and, hence, relatively more S beats are misclassified as normal beats. Another reason is the pattern-wise similarity of the S and N beats. Sometimes it becomes almost impossible to distinguish S-beats from the N-beats even with a trained eye. In some records (e.g., patients 202, 222, 232 and 234) several S beats are present in the sequence, yet only the first S beat displays the timing anomaly, while the others are usually perfectly symmetric but with considerably reduced time intervals. For example, Patient 234 has an episode of junctional tachycardia which last around 25 seconds. It has 50 consecutive beats of supraventricular ectopy (S-beat). The 5 seconds interval of consecutive S beats from this patient's ECG record is shown in Figure 5. This issue arises as the proposed method is limited to beat-by-beat classification. All algorithms that target beat-by-beat classification will eventually suffer in the patient's recording with consecutive S beats. But the proposed method has a superior learning capability and hence the overall performance specially for S beats is much improved compared to earlier "global" ECG classification methodologies in the literature. The results clearly indicate that the proposed approach achieved the top performance in all metrics for N and S beat classification. For V beats, the top performance has been achieved for positive predictivity (Precision) while competitive performance levels have been achieved with the two competing methods for sensitivity and specificity. However, one can note that in [58], [30], and [59], slightly better specificity levels are obtained at the expense of low (<80%) positive predictivity (precision) levels as shown in red in the table. Similarly, [60] and [58] obtain slightly better sensitivity levels; however, their performance levels are quite low in other metrics, including S beat classification (e.g., [58] obtained as low as 30.44% precision level in S-beat classification). The 1D Self-ONN model in [53] achieved the top specificity level in V beat classification; however, it failed in S beat classification too. Such poor performance levels make those competing methods useless and unreliable in clinical practice. Finally, our approach has led to significant improvements in arrhythmia beat detections, especially in S beat classification in all performance metrics over all competing methods. The performance gap sometimes reaches to 30% or even above against [30], [53], [58], [59] and [61].

**Table III:** The confusion matrix represents the results of the beat classification in the MIT-BIH arrhythmia database for 24 records in DS2 (top) and 24 records in DS1 (bottom).

| Ground Truth | Prediction | | |
|---|---|---|---|
| | N | S | V |
| N | 43868 | 266 | 43 |
| S | 269 | 1529 | 36 |
| V | 241 | 36 | 2941 |
| Ground Truth | Prediction | | |
| | N | S | V |
| N | 45355 | 170 | 299 |
| S | 122 | 749 | 72 |
| V | 91 | 6 | 3691 |

**Table IV:** Classification performance of the proposed 1D Self-ONN with Q = 3 and five competing algorithms. The best results are in bold. The performance levels below 80% are shown in red.

| Methods | Acc | Class (N) | | | Class (S) | | | Class (V) | | |
|---|---|---|---|---|---|---|---|---|---|---|
| | | Sen | Ppr | Spe | Sen | Ppr | Spe | Sen | Ppr | Spe |
| **Proposed**[1] | 98.19 | **99.30** | 98.85 | **99.83** | **83.37** | **83.51** | 99.36 | 91.39 | **97.38** | 89.76 |
| **Proposed**[2] | **98.50** | 98.98 | **99.53** | 95.54 | 79.43 | 80.97 | **99.64** | **97.44** | 90.87 | **99.20** |
| Garcia et al [30][1] | 92.4 | 94 | 98 | 82.55 | 61.96 | 52.96 | 97.89 | 87.34 | 59.44 | 95.91 |
| Sellami et al [58][1] | 88.34 | 88.52 | 98.80 | 91.3 | 82.04 | 30.44 | 92.8 | 92.05 | 72.13 | 97.54 |
| Li et al [59][1] | 88.99 | 94.54 | 93.33 | 80.8 | 35.22 | 65.88 | 98.83 | 88.35 | 79.86 | 94.92 |
| Wang et al [60][1] | 97.68 | 99.04 | 98.64 | 87.95 | 70.75 | 77.0 | 99.51 | **94.35** | 95.32 | 99.45 |
| Takalo et al [61][1] | 89.91 | 91.89 | 97 | 76.93 | 62.49 | 55.86 | 98.11 | 89.23 | 50.85 | 94.02 |
| Junaid et al [53][3] | 95.99 | 98.48 | 97.39 | 76.82 | 44.01 | 64.50 | 99.01 | 92.96 | 89.62 | **99.22** |

[1] *Training on DS1 and evaluation on DS2.*
[2] *Training on DS2 and evaluation on DS1.*
[3] *Using the same 1D Self-ONN network as proposed in* [53]*, we evaluated the method for interpatient settings.*

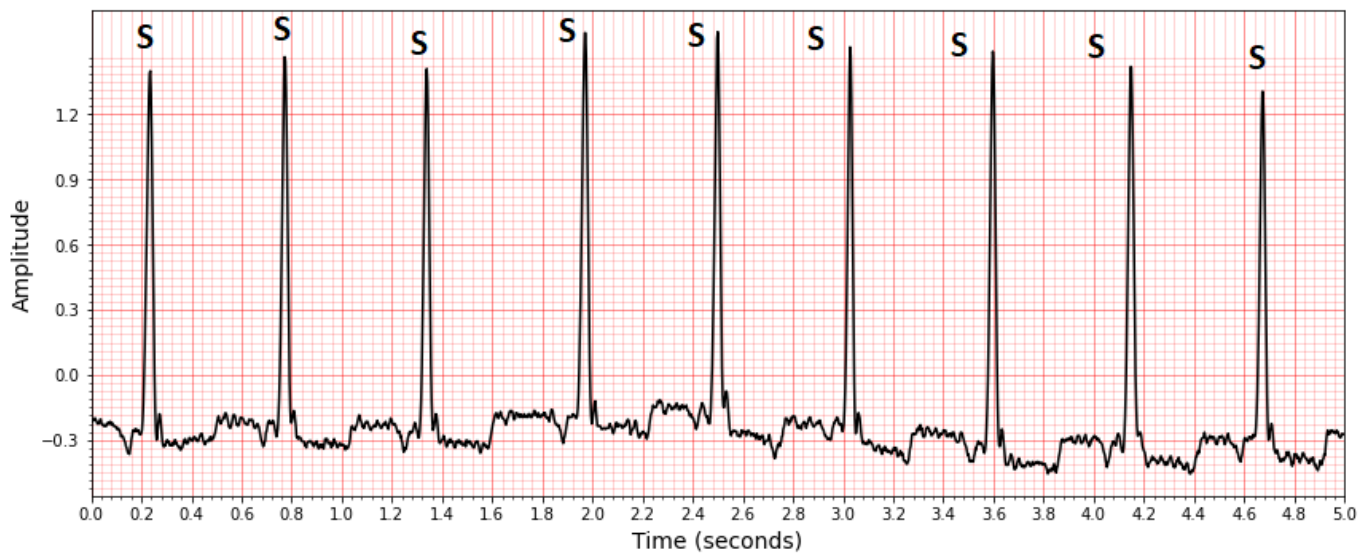

**Figure 5: Five seconds interval from patient 234's ECG record with the ground truth labels.**

In Figure 6, the ROC curves for the proposed method are plotted to show the diagnostic ability of binary classifiers (S Vs Non-S and V Vs Non-V). The ROC curve shows the trade-off between sensitivity (or TPR) and FPR (or 1-specificity). Classifiers that give curves closer to the top-left corner indicate better performance. It is obvious from the ROC curves and area under the curve (AUC) that the proposed system is doing an excellent job in the detection of S and V beats.

*E. Computational Complexity Analysis*

As part of the computation complexity analysis, the total number of layers, total number of neurons, and total number of trainable parameters for each network configuration are calculated and reported in Table V.

As the numbers in the table indicate, along with the compact CNN model proposed in [61], the proposed 1D Self-ONN architecture is the most compact, shallowest network architecture with the least number of parameters. Obviously, the compact CNN model in [61] yields the worst performance level in all metrics shown in Table V. Simultaneously, the proposed method with a similar computational complexity achieves the best performance levels with a significant margin in general.

**Table V: Computational complexity of the networks.**

| Methods | No. of layers | No. of Neurons | No. of trainable parameters |
|---|---|---|---|
| **Proposed** | 4 | 164 | 23,619 |
| Sellami *et al* [58] | 9 | 1344 | 1,395,648 |
| Li *et al*. [59] | 17 | 876 | - |
| Wang *et al*. [60] | 5 | 180 | 26,500 |
| Takalo *et al*. [61] | 4 | 112 | - |

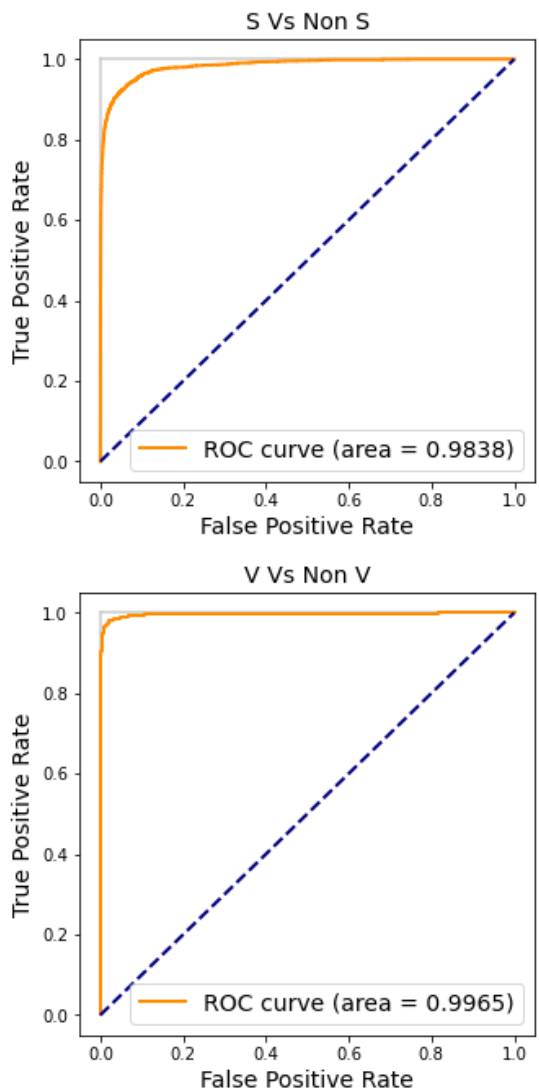

**Figure 6: ROC curves and AUC for S and V beats.**

V. CONCLUSIONS

We have presented a novel approach for classifying heart rhythms from ECG recordings without using any patient-specific data. With the proposed feature injection scheme into the Self-ONN network, our approach exploits both morphological and temporal information of ECG beats to maximize the classification performance. Another critical factor is that each generative neuron in an operational layer is capable of optimizing the nodal operator function of each kernel. Such neuron-level heterogeneity further improves the network diversity and, thus, the learning performance. Finally, with the employed multi-scale signal representation, a high degree of discrimination is accomplished when dealing with normal and arrhythmic ECG signals, especially the S beats. An extensive set of comparative evaluations, performed on the benchmark MIT/BIH arrhythmia database, revealed that our approach outperforms all *state-of-the-art* methods usually with a significant performance gap in SVEB detection. Only the proposed approach can consistently achieve sufficiently high-performance levels required for clinical usage among all the competing methods. Finally, this pioneering study significantly narrows the performance gap between global and *state-of-the-art* patient-specific approaches such as [18] and [48]. Besides the performance superiority, the proposed model is also compact, and thus it can be used in real-time, especially over low-power mobile devices. Due to its highly accurate ECG classification without patient-specific labeled data, our method can serve as an additional diagnostic tool in clinical settings as well as for wearable ECG sensors such as wristbands or smartwatches.

In future work, we plan to further improve the performance and reduce the complexity of the model by exploring improved neuron models in Self-ONNs such as the super (generative) neuron model [62]. Additionally, we intend to expand our research on arrhythmia classification and its generalization to Holter ECGs with low-quality ECG records. In a recent study [63] we showed that the performance of R peak detection drastically decreases when algorithms that are developed for clean ECG signals are applied to noisy and low-quality Holter ECG data. For this purpose, we are planning to use the China Physiological Signal Challenge (2020) database (CPSC-DB) [64], the largest Holter ECG database which contains more than one million beats.

> REPLACE THIS LINE WITH YOUR PAPER IDENTIFICATION NUMBER (DOUBLE-CLICK HERE TO EDIT) <    12[39] S. S. Xu, M. W. Mak, and C. C. Cheung, "Towards End-to-End ECG Classification With Raw Signal Extraction and Deep Neural Networks," *undefined*, vol. 23, no. 4, pp. 1574–1584, Jul. 2019, doi: 10.1109/JBHI.2018.2871510.

[40] U. R. Acharya et al., "A deep convolutional neural network model to classify heartbeats," *Comput Biol Med*, vol. 89, pp. 389–396, Oct. 2017, doi: 10.1016/J.COMPBIOMED.2017.08.022.

[41] U. R. Acharya, H. Fujita, O. S. Lih, Y. Hagiwara, J. H. Tan, and M. Adam, "Automated detection of arrhythmias using different intervals of tachycardia ECG segments with convolutional neural network," *Information Sciences*, vol. 405, pp. 81–90, Sep. 2017, doi: 10.1016/J.INS.2017.04.012.

[42] U. R. Acharya, H. Fujita, O. S. Lih, Y. Hagiwara, J. H. Tan, and M. Adam, "Automated detection of arrhythmias using different intervals of tachycardia ECG segments with convolutional neural network," *Information Sciences*, vol. 405, pp. 81–90, Sep. 2017, doi: 10.1016/J.INS.2017.04.012.

[43] L. Guo, G. Sim, and B. Matuszewski, "Inter-patient ECG classification with convolutional and recurrent neural networks," *Biocybernetics and Biomedical Engineering*, vol. 39, no. 3, pp. 868–879, Jul. 2019, doi: 10.1016/J.BBE.2019.06.001.

[44] S. Kiranyaz, T. Ince, A. Iosifidis, and M. Gabbouj, "Operational neural networks," *Neural Computing and Applications*, vol. 32, no. 11, pp. 6645–6668, Jun. 2020, doi: 10.1007/S00521-020-04780-3.

[45] D. T. Tran, S. Kiranyaz, M. Gabbouj, and A. Iosifidis, "Knowledge Transfer for Face Verification Using Heterogeneous Generalized Operational Perceptrons," *Proceedings - International Conference on Image Processing, ICIP*, vol. 2019-September, pp. 1168–1172, Sep. 2019, doi: 10.1109/ICIP.2019.8804296.

[46] D. T. Tran, S. Kiranyaz, M. Gabbouj, and A. Iosifidis, "Heterogeneous Multilayer Generalized Operational Perceptron," *IEEE Trans Neural Netw Learn Syst*, vol. 31, no. 3, pp. 710–724, Mar. 2020, doi: 10.1109/TNNLS.2019.2914082.

[47] S. Kiranyaz, J. Malik, H. Ben Abdallah, T. Ince, A. Iosifidis, and M. Gabbouj, "Exploiting heterogeneity in operational neural networks by synaptic plasticity," *Neural Computing and Applications*, vol. 33, no. 13, pp. 7997–8015, Jul. 2021, doi: 10.1007/S00521-020-05543-W/FIGURES/18.

[48] S. Kiranyaz, T. Ince, A. Iosifidis, and M. Gabbouj, "Progressive Operational Perceptrons," *Neurocomputing*, vol. 224, pp. 142–154, Feb. 2017, doi: 10.1016/J.NEUCOM.2016.10.044.

[49] S. Kiranyaz, T. Ince, A. Iosifidis, and M. Gabbouj, "Generalized model of biological neural networks: Progressive operational perceptrons," *Proceedings of the International Joint Conference on Neural Networks*, vol. 2017-May, pp. 2477–2485, Jun. 2017, doi: 10.1109/IJCNN.2017.7966157.

[50] S. Kiranyaz, J. Malik, H. ben Abdallah, T. Ince, A. Iosifidis, and M. Gabbouj, "Exploiting heterogeneity in operational neural networks by synaptic plasticity," *Neural Computing and Applications*, vol. 33, no. 13, pp. 7997–8015, Jul. 2021, doi: 10.1007/S00521-020-05543-W/FIGURES/18.

[51] S. Kiranyaz, J. Malik, H. Ben Abdallah, T. Ince, A. Iosifidis, and M. Gabbouj, "Self-Organized Operational Neural Networks with Generative Neurons," *Neural Networks*, vol. 140, pp. 294–308, Apr. 2020, doi: 10.1016/j.neunet.2021.02.028.

[52] J. Malik, S. Kiranyaz, and M. Gabbouj, "Self-Organized Operational Neural Networks for Severe Image Restoration Problems," *Neural Networks*, vol. 135, pp. 201–211, Aug. 2020, doi: 10.1016/j.neunet.2020.12.014.

[53] J. Malik, O. C. Devecioglu, S. Kiranyaz, T. Ince, and M. Gabbouj, "Real-Time Patient-Specific ECG Classification by 1D Self-Operational Neural Networks," *IEEE Transactions on Biomedical Engineering*, 2021, doi: 10.1109/TBME.2021.3135622.

[54] M. Gabbouj et al., "Robust Peak Detection for Holter ECGs by Self-Organized Operational Neural Networks," *IEEE Transactions on Neural Networks and Learning Systems*, pp. 1–12, Mar. 2022, doi: 10.1109/TNNLS.2022.3158867.

[55] "MIT-BIH Noise Stress Test Database v1.0.0." https://www.physionet.org/content/nstdb/1.0.0/ (accessed Dec. 23, 2020).

[56] "MIT-BIH Arrhythmia Database v1.0.0."

[57] "junaidmalik09/fastonn: FastONN - Python based open-source GPU implementation for Operational Neural Networks." https://github.com/junaidmalik09/fastonn (accessed May 15, 2022).

[58] A. Sellami and H. Hwang, "A robust deep convolutional neural network with batch-weighted loss for heartbeat classification," *Expert Systems with Applications*, vol. 122, pp. 75–84, May 2019, doi: 10.1016/J.ESWA.2018.12.037.

[59] Y. Li, R. Qian, and K. Li, "Inter-patient arrhythmia classification with improved deep residual convolutional neural network," *Computer Methods and Programs in Biomedicine*, vol. 214, p. 106582, Feb. 2022, doi: 10.1016/J.CMPB.2021.106582.

[60] T. Wang, C. Lu, Y. Sun, M. Yang, C. Liu, and C. Ou, "Automatic ECG Classification Using Continuous Wavelet Transform and Convolutional Neural Network," *Entropy 2021, Vol. 23, Page 119*, vol. 23, no. 1, p. 119, Jan. 2021, doi: 10.3390/E23010119.

[61] J. Takalo-Mattila, J. Kiljander, and J. P. Soininen, "Inter-patient ECG classification using deep convolutional neural networks," *Proceedings - 21st Euromicro Conference on Digital System Design, DSD 2018*, pp. 421–425, Oct. 2018, doi: 10.1109/DSD.2018.00077.

[62] S. Kiranyaz, J. Malik, M. Yamac, E. Guldogan, T. Ince, and M. Gabbouj, "Super Neurons," Aug. 2021.